**An Open-Source Dual-Loss Embedding Model for Semantic Retrieval in Higher Education**

**Ramteja Sajja**[1,2] (ramteja-sajja@uiowa.edu)
**Yusuf Sermet**[2] (msermet@uiowa.edu)
**Ibrahim Demir**[3,4] (idemir@tulane.edu)

[1] Department of Electrical and Computer Engineering, University of Iowa, Iowa City, USA 52242
[2] IIHR – Hydroscience and Engineering, University of Iowa, Iowa City, USA 52242
[3] River-Coastal Science and Engineering, Tulane University, New Orleans, USA 70118
[4] ByWater Institute, Tulane University, New Orleans, USA 70118

**Corresponding Author**: Ramteja Sajja, ramteja-sajja@uiowa.edu

**ORCID**
Ramteja Sajja: https://orcid.org/0000-0002-2432-2472
Yusuf Sermet: http://orcid.org/0000-0003-1516-8335
Ibrahim Demir: http://orcid.org/0000-0002-0461-1242

**Abstract**
Recent advances in AI have catalyzed the adoption of intelligent educational tools, yet many semantic retrieval systems remain ill-suited to the unique linguistic and structural characteristics of academic content. This study presents two open-source embedding models fine-tuned for educational question answering, particularly in the context of course syllabi. A synthetic dataset of 3,197 sentence pairs, spanning synonymous terminology, paraphrased questions, and implicit-explicit mappings, was constructed through a combination of manual curation and large language model (LLM)-assisted generation. Two training strategies were evaluated: (1) a baseline model fine-tuned using MultipleNegativesRankingLoss (MNRL), and (2) a dual-loss model that combines MNRL with CosineSimilarityLoss to improve both semantic ranking and similarity calibration. Evaluations were conducted on 28 university course syllabi using a fixed set of natural language questions categorized into course, faculty, and teaching assistant information. Results demonstrate that both fine-tuned models outperform strong open-source baselines, including all-MiniLM-L6-v2 and multi-qa-MiniLM-L6-cos-v1, and that the dual-loss model narrows the performance gap with high-performing proprietary embeddings such as OpenAI's text-embedding-3 series. This work contributes reusable, domain-aligned embedding models and provides a replicable framework for educational semantic retrieval, supporting downstream applications such as academic chatbots, retrieval-augmented generation (RAG) systems, and learning management system (LMS) integrations.

# 1. Introduction

The recent proliferation of artificial intelligence (AI) in education has catalyzed a surge in research and deployment of AI-powered tools, underscoring their global relevance and applicability across diverse educational contexts (Bilad et al., 2023; Zahariev et al., 2024; Dulundu, 2024). Innovations such as intelligent tutoring systems, personalized learning platforms, and AI-driven assessment engines are increasingly being integrated into learning management systems (LMS) and institutional workflows (Cano & Troya, 2023; Kumar et al., 2023; Torres et al., 2024). This shift is further exemplified by the emergence of intelligent assistants designed to enhance student engagement and cognitive support. For instance, the Artificial Intelligence-Enabled Intelligent Assistant facilitates adaptive learning pathways, generates quizzes dynamically, and delivers personalized feedback aligned with individual learning profiles (Sajja et al., 2024). These systems rely extensively on natural language interfaces to enable information access, automated support, and seamless interaction, marking a significant departure from conventional keyword-based educational technologies (Yusuf et al., 2024).

This transformation accentuates the critical need for robust and accurate semantic retrieval mechanisms, particularly within structured domains such as academia, where domain-specific terminology, implicit references, and institutional language differ markedly from general-purpose corpora (Rajasurya et al., 2012; Ranjan & Panda, 2021). Recent studies illustrate the growing adoption of conversational AI to deliver cross-disciplinary academic support (Pursnani et al., 2023). These systems leverage advanced document parsing and retrieval techniques to improve learning outcomes across a range of subject areas, including business, environmental science (Samuel et al., 2024; Kadiyala et al., 2024), and quantitative fields (Sajja et al., 2025a). Semantic retrieval models have consistently demonstrated superior precision and recall compared to traditional methods; however, their efficacy remains highly dependent on domain alignment and the extent of linguistic context coverage (Yan et al., 2018; Buscaldi, 2018).

A persistent challenge in existing systems is the limited capacity of general-purpose embedding models to effectively capture academic semantics. Pretrained language models such as BERT, OpenAI embeddings, and other transformer-based architectures are primarily trained on broad, heterogeneous web-scale data. Consequently, these models often yield suboptimal performance when applied to educational question answering (QA) and semantic search tasks without targeted domain adaptation (Tang & Yang, 2024; Alsultan & Razak, 2024; Shen et al., 2024). The limitations are particularly pronounced in low-resource or conceptually complex educational domains, where semantic ambiguity, synonymy, and implicit phrasing are prevalent (Shamdasani et al., 2011; Choudhary et al., 2024; Taipalus, 2023; Dou & Fickas, 2013). These issues are further highlighted by domain-specific implementations, such as AI-driven assistants for certification preparation, which demonstrate the importance of context-sensitive embeddings for achieving high task accuracy and learner alignment (Sajja et al., 2025b).

Moreover, while proprietary models, such as OpenAI's embedding services, exhibit strong performance in many NLP tasks, they introduce critical concerns related to transparency, cost, vendor lock-in, and data governance (Manchanda et al., 2024; Irugalbandara et al., 2023). The black-box nature of such models, combined with restrictive licensing agreements, impedes reproducibility, limits institutional autonomy, and raises barriers to widespread adoption in public education contexts (Zarlenga et al., 2022; Widder et al., 2024). These concerns are especially salient in scenarios requiring explainability and equitable access. Feedback from educators using AI-based learning analytics tools, such as systems employing GPT-4 for engagement tracking and cognitive evaluation, has highlighted both the pedagogical potential and the ongoing concerns surrounding data privacy and the interpretability of AI-generated insights (Sajja et al., 2023b).

Considering these limitations, the research community has increasingly prioritized the development of domain-specific chatbots and embedding models across technical, legal, and medical fields (Sermet and Demir, 2021), demonstrating improved contextual comprehension and downstream task performance (Yunianto et al., 2020; Braun et al., 2021; Mukherjee & Hellendoorn, 2023; Harris et al., 2024; Cruciani et al., 2023). In the educational domain, similar advances, such as AI-augmented virtual teaching assistants that automate responses to course-specific queries, offer promising solutions for reducing cognitive and logistical burdens on both students and instructors (Sajja et al., 2023a). Nevertheless, educational NLP remains comparatively underexplored in the area of embedding development. The absence of embeddings tailored to academic discourse and institutional language continues to constrain the semantic fidelity and utility of AI-powered educational systems.

To address these limitations, this study proposes two open-source embedding models fine-tuned specifically for the educational domain. The models are designed for semantic retrieval tasks grounded in academic discourse, particularly those involving syllabus content and course-related metadata. Such retrieval capabilities are critical components of modern retrieval-augmented generation (RAG) pipelines and enable downstream applications including course-specific chatbots, educational QA systems, and intelligent assistants integrated into learning management systems (LMS). To construct a domain-aligned training corpus, we developed a synthetic dataset of question–answer and synonym pairs using a hybrid approach of manual curation and large language model (LLM)-assisted generation, targeting the linguistic variability and conceptual structures common in academic materials. To explore the impact of training objectives, we trained two variants: one using only the MultipleNegativesRankingLoss (MNRL), and another employing a dual-loss configuration that combines MNRL with CosineSimilarityLoss to enhance both ranking behavior and semantic alignment. The resulting models were benchmarked against proprietary and open-source baselines using real-world university syllabi across multiple departments and institutions. Through this work, we contribute transparent, reusable, and domain-specific embedding models optimized for semantic precision in educational retrieval, advancing the development of trustworthy, performant, and cost-effective academic AI systems.

## 2. Methodology

This section outlines the technical foundation and training strategies used to develop domain-adapted embedding models for educational semantic retrieval. We begin by describing the selection of the base model architecture, followed by the construction of a synthetic dataset tailored to academic QA tasks. The methodology further details two distinct fine-tuning approaches: a contrastive learning strategy using MNRL and a dual-objective framework that integrates MNRL with CosineSimilarityLoss. These components were designed to enhance the model's ability to capture synonymy, paraphrasing, and implicit relationships in educational language, key capabilities for improving retrieval performance in real-world academic applications such as syllabus-based question answering and course bots.

## 2.1. Base Architecture Selection

All fine-tuned models in this study are built upon the *all-MiniLM-L6-v2* architecture, an open-source sentence embedding model released by the Sentence-Transformers project (Reimers& Gurevych, 2019). This model was selected due to its competitive performance on multiple semantic similarity benchmarks, lightweight architecture, and suitability for low-latency inference. These characteristics make it particularly appropriate for educational applications that demand real-time or near-real-time retrieval capabilities, such as course search interfaces, intelligent syllabus assistants, and LMS-integrated question-answering systems. The model's open licensing and availability further supports its adoption in research and production contexts where transparency and adaptability are prioritized.

## 2.2. Dataset Construction and Augmentation

The dataset used for fine-tuning was synthetically constructed through a combination of manual curation and assistance from LLMs, notably GPT-4. The design of the dataset was guided by the need to reflect the nuanced semantic and linguistic characteristics of educational discourse, particularly as observed in course syllabi and academic questions–answering contexts. To support semantic similarity training in this domain, both positive and negative sentence pairs were developed, capturing a broad range of expression patterns and meaning relationships relevant to common academic information needs.

The positive examples, totaling 2,710 pairs, were designed to include a diverse array of semantic phenomena. These encompassed synonymous terminology such as "TA" and "Teaching Assistant" or "Instructor" and "Professor," as well as reformulations that preserve meaning while varying syntactic structure, such as "When are the professor's office hours?" and "What time can I meet the instructor?". Additionally, several examples represented mappings between implicit and explicit expressions, for instance, transforming a question like "Who teaches this class?" into a direct factual statement such as "Instructor: Dr. Smith." The inclusion of such varied positive examples was intended to teach the model to generalize across lexical and syntactic differences commonly encountered in educational documents.

In contrast, 487 negative pairs were constructed to act as difficult semantic distractors. These examples were deliberately crafted to appear structurally similar to positive examples but to be semantically unrelated. Their inclusion was critical for improving the model's capacity to distinguish between true semantic matches and misleadingly similar but incorrect alternatives.

This capability is particularly important in academic contexts, where small semantic deviations can substantially alter meaning.

To ensure comprehensive domain coverage, the dataset was deliberately constructed to include examples spanning course metadata, faculty and instructor information, and teaching assistant references. This design supports generalization across a variety of practical educational queries, enhancing the model's utility in real-world academic applications.

To further improve model generalization and robustness, a series of augmentation techniques were applied. Leveraging LLMs, the dataset was enriched through the generation of paraphrases and linguistic variations of core examples. These augmentations introduced lexical diversity, for example, replacing "When" with "What time", as well as grammatical variation and transformations between question and statement formats, such as rephrasing "What are the professor's office hours?" as "The professor's office hours are at 3 PM." Special attention was given to ambiguous or institution-specific terms, where multiple surface forms exist for the same underlying concept. The central objective of this augmentation phase was to reinforce the model's ability to resolve implicit phrasing into explicit, fact-retrievable statements, an essential skill in semantic retrieval over academic documents.

## 2.3. Fine-Tuning Strategies

To develop embedding models well-suited for capturing the semantic richness and linguistic variability of educational language, we adopted two complementary fine-tuning strategies. Both approaches build upon a shared pretrained base model, *all-MiniLM-L6-v2*, and differ primarily in their choice of training objectives and how supervision is introduced during the optimization process. These strategies aim to improve the model's capacity to support semantic retrieval in academic QA systems, where capturing paraphrase structures, synonymy, and implicit-to-explicit mappings is essential.

### 2.3.1. Contrastive Training with MNRL

The first approach employs the MNRL, a contrastive loss function that optimizes sentence embeddings by bringing semantically related sentence pairs closer together in the embedding space while pushing unrelated pairs further apart. This loss function operates at the batch level: for each anchor-positive pair in the batch, all other batch elements act as implicit negatives. This formulation enables efficient learning even in the absence of explicitly annotated negative samples for every pair. Training is conducted using a single DataLoader, populated exclusively with positive sentence pairs, typically synonymous or paraphrased QA pairs. Negative examples are implicitly constructed through contrastive sampling within each batch. This setup is highly effective for retrieval-focused tasks where ranking semantically similar results is the primary objective.

### 2.3.2. Dual-Objective Training with MNRL and Cosine Similarity

The second approach extends the optimization objective by incorporating an additional CosineSimilarityLoss alongside MNRL. While MNRL encourages correct relative ordering of similarity among embeddings, CosineSimilarityLoss provides explicit supervision of similarity scores, directly aligning embedding distances with semantic labels. This is especially useful in cases where small variations in language can carry significant meaning in academic contexts. To facilitate this joint optimization, the training pipeline is structured around two separate DataLoaders: (i) One DataLoader supplies positive-only pairs used with MNRL. (ii) The second DataLoader includes both positive and negative sentence pairs, each explicitly labeled: 1 for semantically similar pairs and 0 for dissimilar pairs, used for CosineSimilarityLoss.

During training, batches are alternated or sampled from both DataLoaders, and losses from both objectives are computed and combined in a multi-objective fine-tuning loop. This dual-loss approach enables the model to learn both robust ranking behavior and calibrated similarity scoring, improving its performance on downstream tasks involving question matching and semantic retrieval in educational settings.

### 2.4. Training Configuration and Optimization Setup

The fine-tuning experiments were conducted under two distinct training setups, each aligned with its respective optimization strategy. For the single-loss approach (MNRL-only), the model was trained for 25 epochs using the MultipleNegativesRankingLoss objective. A batch size of 64 was used, with 15% of total steps allocated for warmup. The AdamW optimizer was applied with a learning rate of $2e^{-5}$, alongside a WarmupCosine learning rate scheduler. Weight decay was set to 0.01 to regularize the model and mitigate overfitting. Mixed-precision (AMP) was disabled, as all training was conducted on CPU.

For the dual-loss approach, which jointly optimized both MNRL and CosineSimilarityLoss, the model was similarly trained for 25 epochs, but with a lower learning rate of $1e^{-5}$ and a slightly reduced warmup proportion of 10%. The batch size was maintained at 64, with the same WarmupCosine scheduler and AdamW optimizer configuration. Weight decay remained at 0.01. To accommodate the dual-objective setup, separate DataLoaders were used for each loss, and training alternated between them in a coordinated multi-objective loop.

### 3. Results & Discussions

This section presents the evaluation results and accompanying analysis of the proposed fine-tuned embedding models within an educational semantic retrieval setting. We compare our models, trained using contrastive and dual-loss strategies, against both proprietary and widely adopted open-source baselines. The goal is to assess their ability to retrieve contextually relevant information from real-world university syllabi in response to natural language questions, a key requirement for educational QA systems and RAG-based pipelines.

We first outline the experimental setup, including the chunk-based retrieval method and evaluation corpus. We then report model accuracy across three categories of academic metadata and analyze retrieval performance across embedding models. Finally, we discuss the semantic benefits of dual-loss optimization and the broader implications for developing open-source alternatives to commercial embedding systems in higher education contexts.

## 3.1. Evaluation Framework

To evaluate retrieval performance in realistic academic settings, we adopt a task-specific benchmark using manually validated natural language queries over a diverse corpus of university syllabi. This framework is designed to test how well each embedding model can support semantic similarity tasks relevant to educational applications. We describe the chunking approach, the retrieval pipeline, the question design, and the benchmark models used for comparison.

### 3.1.1. Embedding-Based Retrieval with Cosine Similarity

To evaluate the semantic retrieval capabilities of embedding models in an educational context, we adopt a chunk-based retrieval framework. Each syllabus document is segmented into fixed-size text chunks of approximately 300 words or fewer, and embeddings are precomputed for each chunk using the embedding model under evaluation. The evaluation is performed over 28 diverse university syllabus files, spanning multiple institutions and a range of departments including computer science, anthropology, psychology, biology, education, and engineering. This corpus reflects a realistic variety in structure, terminology, and writing style found in real-world educational materials. A fixed set of natural language questions is used to test retrieval performance, grouped into three categories: *Course Information*, *Faculty Information*, and *Teaching Assistant Information*. Each category includes multiple phrasings to assess model robustness to linguistic variation, as shown in Table 1.

Table 1: Evaluation Questions Categorized by Educational Metadata Type

| Category | Natural Language Questions |
|---|---|
| Course Information | - How many credit hours is this course worth? |
|  | - What are the semester hours for this course? |
|  | - What is the credit hour value of this course? |
| Faculty Information | - What is the name of the instructor? |
|  | - What is the professor's name? |
|  | - What is the lecturer's name? |
| Teaching Assistant Information | - What is the name of the TA? |
|  | - What are the TA's name? |
|  | - What is the name of the Teaching Assistant? |
|  | - What are the Teaching Assistant's name? |

For each question, an embedding is generated using the same model. Cosine similarity is then computed between the question embedding and all chunk embeddings from a given syllabus document (Gunawan et al., 2018). Three most similar chunks are retrieved and concatenated to form a context string. This context is passed to a lightweight generative model (GPT-4o-mini), which produces a natural language answer. The final answer is evaluated manually against the full syllabus document to determine whether it is factually supported by the original content.

### 3.1.2. Benchmarked Models

To evaluate the effectiveness of our fine-tuned embedding models in educational semantic retrieval, we benchmark against both proprietary and open-source baselines as shown in Table 2.

Table 2: Overview of Evaluated Embedding Models

| Category | Model Name | Description |
|---|---|---|
| **Proprietary (Closed)** | text-embedding-ada-002 (Greene et al., 2022) | OpenAI's widely adopted embedding model with strong general performance. |
| | text-embedding-3-small (OpenAI, 2024) | New generation embedding model with improved quality-to-cost ratio. |
| | text-embedding-3-large (OpenAI, 2024) | Most advanced in OpenAI's embedding lineup, optimized for performance. |
| **Open Source** | all-MiniLM-L6-v2 (Reimers& Gurevych, 2019) | Popular, efficient model for general-purpose semantic similarity tasks. |
| | multi-qa-MiniLM-L6-cos-v1 (Thakur et al., 2021) | Fine-tuned for multi-domain QA with cosine-based retrieval optimization. |
| | Msmarco-distilbert-base-v4 (Reimers& Gurevych, 2020) | Trained on MS MARCO for passage retrieval and QA applications. |
| | nli-roberta-base-v2 (Reimers& Gurevych, 2019) | Fine-tuned on natural language inference tasks, adapted for semantic alignment. |
| **Ours (Educational)** | Fine-tuned Educational Model (MNRL) | Fine-tuned on curated educational QA pairs using MultipleNegativesRankingLoss. |
| | Fine-tuned Educational Model (Dual-loss) | Jointly optimized using ranking and similarity objectives for deeper semantics. |

### 3.1.3. Benchmark Design

To systematically evaluate the retrieval quality of different embedding models in academic contexts, we employ a category-specific accuracy benchmark. The benchmark measures whether a model successfully retrieves supporting context that leads to a correct answer when passed through a lightweight generative model.

Evaluation is conducted across three core categories of educational metadata that are commonly encountered in academic question-answering tasks. The first category, Course Information, includes queries related to credit hours, course duration, and other curriculum-related attributes. The second category, Faculty Information, focuses on identifying the instructor or professor associated with the course. The third category, Teaching Assistant Information, targets queries regarding the presence and identity of teaching assistants. These categories were selected to reflect distinct but frequently queried types of academic metadata, providing a structured framework for analyzing the semantic retrieval capabilities of different embedding models.

For each embedding model, the same fixed pool of 28 syllabus documents and the same natural language question set (outlined in Table 1) are used to ensure consistency and comparability. The output from the generative model (GPT-4o-mini), given the top-3 retrieved chunks, is manually evaluated for factual correctness by checking it against the original syllabus content. A binary accuracy metric is used, where each answer is labeled as either *valid* (i.e., the answer is explicitly supported by the syllabus file) or *invalid*. This evaluation setup enables a direct comparison of retrieval effectiveness across open-source, proprietary, and fine-tuned embedding models, while highlighting performance variation across different types of educational queries.

### 3.2. Quantitative Accuracy

To assess the retrieval effectiveness of each embedding model, we measure accuracy across three evaluation categories: Course Information, Faculty Information, and Teaching Assistant Information. The metric used is binary accuracy, determined by whether the generated answer is factually supported by the syllabus content.

**Table 3**: Accuracy Comparison Across Embedding Models and Evaluation Categories

| | Model | Course Information | Faculty Information | TA Information |
|---|---|---|---|---|
| CLOSED | text-embedding-3-large | 100.00% | 95.24% | 90.18% |
| | text-embedding-3-small | 100.00% | 89.29% | 77.68% |
| | text-embedding-ada-002 | 100.00% | 91.67% | 83.93% |
| OURS | **Fine-tuned Educational Model (MNRL)** | **100.00%** | **84.52%** | **87.50%** |
| | **Fine-tuned Educational Model (Dual-loss)** | **100.00%** | **88.10%** | **87.50%** |
| OPEN SOURCED | all-MiniLM-L6-v2 | 92.86% | 73.81% | 69.64% |
| | msmarco-distilbert-base-v4 | 98.81% | 79.76% | 77.68% |
| | multi-qa-MiniLM-L6-cos-v1 | 97.62% | 85.71% | 71.43% |
| | nli-roberta-base-v2 | 97.62% | 85.71% | 83.93% |

The results presented in Table 3 demonstrate a clear advantage of fine-tuning embedding models specifically for the educational domain. The dual-loss fine-tuned model, which jointly optimizes MultipleNegativesRankingLoss and CosineSimilarityLoss, outperforms all open-source baselines across all three categories and comes remarkably close to the performance levels of proprietary OpenAI models. In the Teaching Assistant Information category, which requires fine-grained semantic understanding and synonym resolution, the dual-loss model achieves 87.50% accuracy, matching the performance of the MNRL-only model and narrowing the gap with OpenAI's text-embedding-3-large model (90.18%).

The Course Information category shows high performance across nearly all models, with both fine-tuned models and OpenAI embeddings achieving 100% accuracy. Notably, some syllabus files did not explicitly contain semester or credit hour information; in those instances, an answer of *"Sorry, I don't know"* was treated as a valid response if it accurately reflected the absence of relevant content in the syllabus. This approach emphasizes the importance of accurate retrieval, not only producing correct answers but also knowing when not to hallucinate responses in data-scarce contexts.

In the Faculty Information category, the dual-loss model shows a marked improvement over the MNRL-only model (88.10% vs. 84.52%), and even surpasses several OpenAI baselines like text-embedding-3-small (89.29%) and text-embedding-ada-002 (91.67%). This performance underscores the dual-loss model's ability to better align embedding distances with subtle semantic distinctions in educational language.

Overall, the findings affirm that domain-specific fine-tuning, especially using a multi-objective loss strategy, can significantly enhance semantic retrieval in structured academic domains. The dual-loss model not only bridges the performance gap with closed-source solutions but also advances open-source alternatives that are lower-cost, transparent, and adaptable for educational institutions.

## 3.3. Discussion

This section interprets the empirical findings presented in the evaluation and examines their implications for embedding model development in educational contexts. We focus on two central themes: the performance impact of using a dual-loss training strategy, and the broader need for improved open-source embedding models tailored to specialized domains like education. The analysis emphasizes how targeted fine-tuning, combined with semantically rich supervision, enhances the retrieval of academically relevant content and narrows the performance gap between open and proprietary systems. We also explore how our methodology addresses the limitations of generic embeddings when applied to structured educational queries and documents such as syllabi. Together, these insights inform ongoing efforts to build more transparent, adaptable, and domain-aligned semantic retrieval systems for higher education.

### 3.3.1. Dual-Loss Training Yields Better Semantic Coverage

Our results demonstrate that combining MNRL and CosineSimilarityLoss offers a measurable benefit over using MNRL alone, especially in domain-specific tasks where subtle distinctions in meaning are critical. MNRL excels at enforcing relative similarity, ensuring that a positive pair is more similar than other items in a batch, but it lacks direct supervision over the magnitude of similarity scores. This limitation can be problematic in educational QA tasks where numerous phrases may share similar surface structures but differ semantically. For instance, the model must distinguish between "office hours" and "lecture times," or between a professor's name and a teaching assistant's name, even when both are presented in similar textual forms.

By integrating CosineSimilarityLoss, we explicitly guide the model to align its similarity scores with binary semantic labels (1 for similar, 0 for dissimilar). This adds a layer of global consistency in embedding space geometry, reinforcing both ranking behavior and semantic calibration. The use of dual DataLoaders, one supplying positive-only pairs for MNRL and another providing labeled pairs for CosineSimilarityLoss, enables flexible sampling and efficient optimization. Together, this dual-loss formulation leads to more robust embeddings, capable of generalizing across linguistic variations and handling ambiguous queries with greater accuracy.

The improvements were particularly evident in categories like Teaching Assistant Information, where lexical overlaps and institutional jargon frequently cause confusion in less specialized models. The dual-loss model reduced such misclassifications, demonstrating a stronger grasp of domain-relevant terminology and context. This suggests that the approach not only improves retrieval performance but also enhances the semantic fidelity of the embedding space, a critical requirement for downstream educational applications like course chatbots, LMS assistants, and syllabus-based question answering.

### 3.3.2. Need for Better Open-Source Embedding Models

While domain-specific fine-tuning significantly improved performance, our results also underscore the persistent performance gap between open-source and proprietary models. OpenAI's embeddings, particularly text-embedding-3-large, remain the top performers across all evaluation categories, achieving the highest accuracy in Faculty and Teaching Assistant information retrieval. However, such models are closed-source, reliant on paid API access, and raise significant concerns related to data privacy, vendor lock-in, and lack of customization. These are especially critical limitations for academic institutions that handle sensitive student and course data, or that seek long-term control over their infrastructure.

Our fine-tuned dual-loss model not only closes much of this gap but does so using entirely open-source tools and models. This result highlights the untapped potential of community-maintained architectures, such as all-MiniLM-L6-v2, coupled with task-specific data curation and training strategies. It suggests that, while general-purpose models trained on broad internet corpora are powerful, they are insufficient for high-stakes academic tasks without targeted adaptation.

Moreover, our findings reinforce the importance of domain alignment in semantic retrieval. The educational domain exhibits specialized terminology, non-standard phrasing, and implicit document structures (e.g., tabular or formatted lists in syllabi), which are often poorly represented in general-purpose corpora. Fine-tuning on a dataset that mirrors this linguistic and structural variation results in embeddings that are better aligned with the practical needs of educational QA systems.

This work advocates for a more systematic effort to build and release domain-specific embedding models that prioritize transparency, replicability, and cost-efficiency, and core values for the academic community. It also underscores the value of synthetic but targeted training datasets, which, even in small volumes, can significantly elevate the quality of retrieval for niche domains.

## 4. Conclusion & Future Work

This study introduces two open-source embedding models fine-tuned specifically for semantic retrieval tasks in the educational domain. Built upon the all-MiniLM-L6-v2 architecture, the models were trained on a synthetically constructed dataset comprising question–answer and synonym pairs, generated through a combination of manual curation and LLM-assisted augmentation. The dataset captures a broad spectrum of academic semantic phenomena, including synonymous terminology, paraphrased queries, and implicit-explicit mappings frequently observed in educational texts such as syllabi.

To evaluate the effect of fine-tuning objectives on model performance, two distinct training strategies were implemented. The first approach employed a standard contrastive learning setup using MNRL, which encourages the model to distinguish between semantically similar and dissimilar pairs. The second approach introduced a dual-loss framework that combines MNRL with CosineSimilarityLoss, thereby incorporating both relative ranking and explicit similarity supervision. This dual-objective strategy was designed to better capture subtle distinctions in academic language and improve alignment between embedding distance and semantic intent.

Empirical evaluation was conducted using a chunk-based retrieval framework across 28 university syllabi from diverse institutions and departments. Each model was assessed using a fixed set of natural language questions grouped into three categories: course information, faculty information, and teaching assistant information. Manual accuracy evaluation revealed that both fine-tuned models outperformed several open-source baselines, including multi-qa-MiniLM-L6-cos-v1, msmarco-distilbert-base-v4, nli-roberta-base-v2, and the base all-MiniLM-L6-v2. Notably, the dual-loss model consistently approached the performance of proprietary OpenAI embeddings, highlighting the potential of targeted fine-tuning for narrowing the performance gap between open and closed systems.

These results affirm the viability of domain-specific fine-tuning in improving semantic retrieval for educational applications, with implications for question-answering pipelines, academic chatbots, and LMS integrations. However, several limitations remain that warrant further investigation. One key limitation is the chunking strategy employed during evaluation. Current fixed-size segmentation methods can split semantically related content across multiple chunks or include irrelevant text within a single unit, introducing noise and limiting retrieval precision. Future work should explore more intelligent chunking approaches, such as segmenting based on document structure, semantic boundaries, or discourse coherence, to better preserve meaningful content units.

Additionally, embedding quality may be compromised by the presence of boilerplate or low-information content, such as institutional policies or footer text, that dilutes the semantic signal. Preprocessing steps that filter or down-weight such content could enhance the effectiveness of chunk embeddings. Furthermore, incorporating harder negative examples during training and adopting curriculum learning strategies may further improve the model's discriminative capabilities, especially in cases involving closely related distractor content. Finally, while this study focused on syllabus documents, future research should consider extending the evaluation to other educational materials such as lecture slides, academic policies, and program descriptions. Expanding the domain coverage would enhance model robustness and facilitate broader adoption across educational QA systems.

In contrast to proprietary commercial solutions, this work supports ethical, transparent, and reproducible AI development for higher education by releasing both training methodology and fine-tuned models under an open license. Our findings show that domain-specific fine-tuning, especially when supported by LLM-generated supervision and dual-loss optimization, can deliver high-performing educational embedding models that rival commercial alternatives. Future work will explore smarter chunking strategies, better filtering of noisy content, and application to broader educational artifacts such as academic policies, lecture notes, and institutional guidelines. This approach lays a foundation for LLM-augmented educational QA systems, LMS integration, and scalable personalization in AI-supported teaching and learning.

**Declaration of Generative AI and AI-Assisted Technologies**
During the preparation of this manuscript, the authors used ChatGPT, based on the GPT-4o model, to improve the flow of the text, correct grammatical errors, and enhance the clarity of the writing. The language model was not used to generate content, citations, or verify facts. After using this tool, the authors thoroughly reviewed and edited the content to ensure accuracy, validity, and originality, and take full responsibility for the final version of the manuscript.

**Funding**
Funding for this project was provided by the University of Iowa's Innovations in Teaching with Technology Awards.

**Availability of data and materials**
The training dataset generated during this study is publicly available at: https://github.com/uihilab/educational-qa-embeddings.
The fine-tuned embedding models are released as open-source resources and can be accessed on Hugging Face at the following links:
- Dual-loss fine-tuned model: https://huggingface.co/rsajja/Fine-tuned-Educational-Model-Dual-Loss
- MNRL-only fine-tuned model: https://huggingface.co/rsajja/Fine-tuned-Educational-Model-MNRL

**Competing Interests**
The authors declare that they have no competing interests.

**CRediT Author Statement**
**Ramteja Sajja**: Conceptualization, Methodology, Software, Validation, Formal analysis, Investigation, Data Curation and Writing - Original Draft. **Yusuf Sermet**: Writing - Review & Editing, Project administration, Validation, and Funding acquisition. **Ibrahim Demir**: Writing - Review & Editing, Supervision, Funding acquisition, and Resources.